\documentclass{article}

\usepackage[main, final]{neurips_2026}
\usepackage[utf8]{inputenc} % allow utf-8 input
\usepackage[T1]{fontenc}    % use 8-bit T1 fonts
\usepackage{hyperref}       % hyperlinks
\usepackage{url}            % simple URL typesetting
\usepackage{booktabs}       % professional-quality tables
\usepackage{amsfonts}       % blackboard math symbols
\usepackage{nicefrac}       % compact symbols for 1/2, etc.
\usepackage{microtype}      % microtypography
\usepackage{xcolor}         % colors
\usepackage{graphicx}
\usepackage{amsmath}
\usepackage{subcaption}
\usepackage{multirow}
\usepackage{color}
\usepackage{colortbl}
\usepackage{wrapfig}
\definecolor{tableBlue}{rgb}{0.424,0.557,0.749}
\definecolor{lightblue}{HTML}{dfebf7}
\newcommand{\tableLineColor}{\rowcolor{lightblue}}
\usepackage{booktabs}
\usepackage{arydshln} % 必须在所有表格宏包之后加载

\title{WaveletLoRA: Frequency-Aware Content-Style Decomposition for Personalized Image Generation}

\author{%
  Peiyao Wang\textsuperscript{1,2}\thanks{Equal Contribution.}
  \And
  Jiahui Sun\textsuperscript{1,2}\footnotemark[1]
  \And
  Weining Wang\textsuperscript{1,2}\thanks{Corresponding Author.}
  \And
  Jing Liu\textsuperscript{1,2}
}

\begin{document}

\maketitle

\begin{center}
\vspace{-4ex}
\textsuperscript{1}Institute of Automation, Chinese Academy of Sciences\\
\textsuperscript{2}School of Artificial Intelligence, University of Chinese Academy of Sciences
\vspace{2ex}
\end{center}

% \author{%
%   David S.~Hippocampus\thanks{Use footnote for providing further information
%     about author (webpage, alternative address)---\emph{not} for acknowledging
%     funding agencies.} \\
%   Department of Computer Science\\
%   Cranberry-Lemon University\\
%   Pittsburgh, PA 15213 \\
%   \texttt{hippo@cs.cranberry-lemon.edu} \\
%   % examples of more authors
%   % \And
%   % Coauthor \\
%   % Affiliation \\
%   % Address \\
%   % \texttt{email} \\
%   % \AND
%   % Coauthor \\
%   % Affiliation \\
%   % Address \\
%   % \texttt{email} \\
%   % \And
%   % Coauthor \\
%   % Affiliation \\
%   % Address \\
%   % \texttt{email} \\
%   % \And
%   % Coauthor \\
%   % Affiliation \\
%   % Address \\
%   % \texttt{email} \\
% }

\begin{abstract}
%Extracting and cleanly separating content and style from a single image is a fundamental challenge for personalized image generation.
%Existing methods primarily operate in the spatial domain and do not explicitly exploit the natural frequency separation of images, where low-frequency components encode global style and high-frequency details capture structural content.
%In this paper, we introduce WaveletLoRA, a frequency‑aware framework that decomposes the input image into sub-bands containing different frequency domain components using the discrete wavelet transform, and trains a dedicated LoRA branch for each sub-band.
%The low‑frequency branch serves as a style LoRA, while the high‑frequency branches are dynamically fused through a dense Mixture‑of‑Experts module to form a content LoRA.
%A weight‑level frequency disentanglement loss and an AdaIN‑based style‑statistics alignment term further enforce clean separation.
%To systematically evaluate decomposition quality, we construct WaveBench, a large‑scale dataset of 1{,}000 images with meticulously curated prompts, and design a VLM‑based evaluation protocol that aligns more consistently with human judgment than conventional feature‑space metrics.
%Experiments on both U‑Net and DiT architectures show that WaveletLoRA achieves state‑of‑the‑art content–style decomposition and enables flexible cross‑image composition with low-cost fine-tuning.
Content-style decomposition from a single image is a fundamental challenge in personalized image generation, aiming to separate subject identity from visual style for flexible recomposition. Existing approaches primarily operate in the spatial domain, without explicitly exploiting the distinct frequency characteristics naturally associated with content and style. In this paper, we propose WaveletLoRA, a novel frequency-aware framework for explicit content-style decomposition through wavelet-domain supervision. Specifically, we first apply a multi-level discrete wavelet transform (DWT) to decompose the reference image into low-frequency and high-frequency subbands. We then assign a dedicated LoRA branch to each subband, where the low-frequency branch models global appearance for style acquisition, while the high-frequency branches capture structural details for content preservation. To further improve disentanglement quality, we introduce a dense Mixture-of-Experts aggregation module together with a frequency-domain regularization objective, which reduces style leakage and enhance content fidelity. In addition, we establish WaveBench and develop a VLM-based evaluation protocol that explicitly measures both attribute similarity and attribute separation for disentanglement assessment. Extensive experiments demonstrate that WaveletLoRA consistently outperforms existing methods and produces results that better align with human preference.
%To further improve disentanglement quality, we introduce a dense Mixture-of-Experts (MoE) module that dynamically aggregates high-frequency representations into a unified content representation.
%In addition, we design a frequency-domain disentanglement objective consisting of a weight-level frequency regularization term and an AdaIN-based style statistics alignment loss to encourage clean separation between content and style.
Code and the proposed dataset will be released in \url{https://github.com/PinkYearwww/WaveletLoRA}.
%To facilitate rigorous evaluation, we construct WaveBench, a comprehensive benchmark comprising 1{,}000 curated examples, together with a VLM-based evaluation protocol that exhibits stronger correlation with human preference than conventional feature-space metrics.
%Extensive experiments demonstrate that WaveletLoRA achieves state-of-the-art performance in content preservation and style fidelity under a parameter-efficient fine-tuning setting.
%Extensive experiments across both U-Net and Diffusion Transformer (DiT)-based architectures demonstrate that WaveletLoRA achieves state-of-the-art performance in content preservation and style fidelity under a parameter-efficient fine-tuning setting.
\end{abstract}

\begin{figure}[h!]
\vspace{-1ex}
    \centering
    \includegraphics[width=\textwidth]{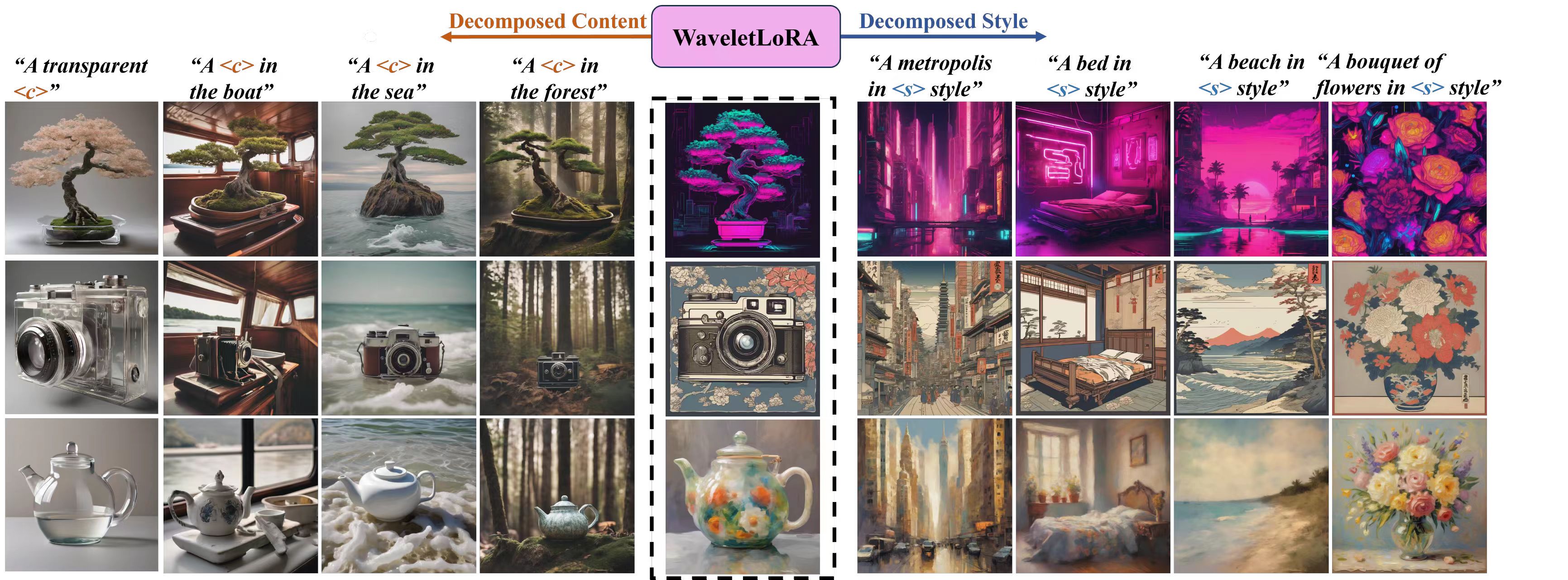}    
    \caption{Given a single input image (middle), WaveletLoRA learns a disentangled content or subject LoRA (left) and style LoRA (right) that can be used to generate new images with the learned concepts.}
    \label{fig:waveletlora}
    \vspace{-4px}
\end{figure}

\section{Introduction}
\label{sec:intro}

Personalized image generation has recently emerged as a major research direction in generative modeling, driven by the growing demand for controllable and customized visual content generation ~\cite{ruiz2023dreambooth, shah2024ziplora, sohn2023styledrop, frenkel2024implicit, wu2025uso}. Although modern text-to-image diffusion models have demonstrated remarkable generation capability, textual prompts alone often fail to specify fine-grained instance-level attributes, such as the identity of a particular subject or the unique visual characteristics of a specific artistic style.
To address this limitation, example-driven personalization methods, such as DreamBooth~\cite{ruiz2023dreambooth} and StyleDrop~\cite{sohn2023styledrop}, fine-tune a pretrained model on a small set of reference images to inject subject or style concepts into the generation process. While effective, these methods typically specialize in either content or style in isolation. Disentangling both factors simultaneously from a single image, while preserving clean separation and recompositional flexibility, remains a challenging open problem.

To address single image content–style decomposition, several methods have recently been proposed~\cite{frenkel2024implicit, shah2024ziplora, liu2025unziplora}. 
%However, existing methods predominantly operate in the spatial or feature domain and rely on architectural constraints or optimization heuristics to separate content and style, without explicitly modeling their intrinsic frequency characteristics.
However, existing methods predominantly operate in the spatial or feature domain and learn to separate content and style through architectural constraints and optimization heuristics, rather than explicitly modeling the intrinsic and fundamental disparities between structural content and visual appearance.
As a result, content and style representations often remain entangled, leading to style leakage in content generation or subject residue in style transfer.
We argue that a key reason for this limitation is the neglect of an important signal-processing prior: content and style naturally exhibit different frequency characteristics.

Classical studies in image processing have shown that high-frequency components primarily encode structural information, such as edges and fine-grained textures, whereas low-frequency components mainly capture global appearance attributes, including color distribution and artistic style  ~\cite{gatys2016image, wang2018high}. This observation suggests that frequency decomposition provides a natural inductive bias for disentangling content and style. Existing frequency-aware generative methods have largely relied on Fourier-domain analysis~\cite{tancik2020fourier, durall2020watch, wang2022fregan, karras2024analyzing}, which captures global spectral statistics but lacks spatial locality. In contrast, the discrete wavelet transform (DWT) provides localized multi-scale frequency decomposition while preserving both spatial and spectral information, making it more suitable for separating structural content from stylistic appearance.

Motivated by these insights, we propose \textbf{WaveletLoRA}, a novel frequency-aware framework for personalized content–style decomposition.
Specifically, we first decompose the reference image into low-frequency and high-frequency wavelet subbands, and assign dedicated LoRA branches to each subband for explicit frequency-domain supervision. The low-frequency branch is optimized to model global style characteristics, while the high-frequency branches focus on preserving multi-scale structural details.
To effectively integrate structural information across different frequency bands,  we further introduce a dense Mixture-of-Experts (MoE) module that dynamically aggregates high-frequency LoRA branches into a unified content representation. In addition, we design a frequency-domain disentanglement objective that combines weight-level frequency regularization with AdaIN-based style statistics alignment to further encourage clean separation between content and style.
%that disentangles content and style by supervising LoRA modules with frequency subbands obtained via wavelet decomposition. By explicitly separating low-frequency style cues from high-frequency content details, WaveletLoRA achieves clean decomposition without architectural restrictions.
%Specifically, we first perform a multi-level DWT on the reference image to obtain one low-frequency subband and multiple high-frequency subbands. We then assign an independent LoRA branch to each subband, enabling explicit frequency-domain supervision during training. The low-frequency branch is optimized to model global style characteristics, while the high-frequency branches focus on preserving multi-scale structural details. To effectively aggregate structural information across different frequency bands, we further introduce a dense Mixture-of-Experts (MoE) module that dynamically fuses high-frequency LoRA branches into a unified content representation. In addition, we design a frequency-domain disentanglement objective that combines weight-level frequency regularization with AdaIN-based style statistics alignment to further encourage clean separation between content and style.
Another challenge in single-image content–style decomposition lies in evaluation. Existing metrics, such as DINOv2~\cite{hertz2022prompt}  and CLIP-I~\cite{cai2025z}, primarily measure holistic visual similarity and cannot reliably assess disentanglement quality. To address this issue, we construct WaveBench, a large-scale benchmark together with a VLM-based evaluation framework that explicitly measures both attribute similarity and attribute separation.

In summary, our main contributions are as follows:
\begin{itemize}
\item We observe that content and style exhibit distinct frequency characteristics, and propose wavelet-domain decomposition as a principled prior for single-image content–style disentanglement.
\item We propose WaveletLoRA, a novel frequency-aware framework that learns disentangled content and style LoRAs from multi-scale wavelet subbands, together with dense MoE aggregation and frequency-domain regularization for improved structural preservation and reduced style leakage.
\item We establish WaveBench, a large-scale benchmark for single-image content–style decomposition with a VLM-based evaluation protocol, and demonstrate state-of-the-art performance in content preservation and style fidelity.
% \item We propose WaveletLoRA, a frequency-aware framework that, for the first time, introduces the discrete wavelet transform into personalized content–style disentanglement, leveraging multi‑scale subband supervision to naturally separate low‑frequency style from high‑frequency content. We further validate WaveletLoRA on both U-Net and DiT diffusion backbones, demonstrating strong generalization and state-of-the-art disentanglement quality.
%\item We propose WaveletLoRA, a novel frequency-aware framework that integrates discrete wavelet transforms into personalized disentanglement. By leveraging multi-scale subband supervision, it decouples low-frequency style from high-frequency content across both U-Net and DiT backbones, achieving state-of-the-art quality and generalization.
%\item We establish WaveBench, a new benchmark for single-image content–style decomposition,  and perform comprehensive evaluations using both standard feature-space metrics and VLM-based assessments.
%\item Extensive experiments on both U-Net and DiT diffusion architectures demonstrate that WaveletLoRA achieves state-of-the-art performance in content preservation, style fidelity, and disentanglement quality under a parameter-efficient fine-tuning regime.
\end{itemize}

\section{Related Works}

\textbf{Personalized image generation.}
%Custom image generation to produce specific concepts has been a central pursuit in generative modeling. Image stylization, studied for over two decades, achieved early breakthroughs with convolutional neural networks~\cite{gatys2016image, johnson2016perceptual, huang2017arbitrary, li2017universal, park2019arbitrary} and later advanced through GAN-based one-shot stylization~\cite{chong2022jojogan, gal2022stylegan, kwon2023one, liu2021blendgan, ojha2021few, shah2022multistylegan, sohn2023styledrop, wang2022ctlgan, yang2023one, yang2022pastiche, zhu2021mind}, yet these methods typically transfer a given style onto a separate content image and are confined to single domains without text-based control. 
With the rise of diffusion models~\cite{ho2020denoising, rombach2022high, song2020denoising}, researchers have developed powerful fine-tuning strategies tailored to personalized image generation. DreamBooth~\cite{ruiz2023dreambooth} and Textual Inversion~\cite{gal2022image} enabled faithful subject or style learning from a handful of reference images.
% while StyleDrop~\cite{sohn2023styledrop} further raised style fidelity through iterative refinement. 
However, these methods~\cite{sohn2023styledrop,gal2022image,ruiz2023dreambooth} are predominantly designed to capture a single concept at a time. To address the more challenging scenario of extracting multiple concepts from one image, Inspiration Tree~\cite{vinker2023concept} and CusConcept~\cite{xu2025cusconcept} learn hierarchical or customized text embeddings via textual inversion, but the resulting embeddings offer limited generation control compared to full weight-space adaptations. 
% ConceptExpress~\cite{hao2024conceptexpress} and Break-A-Scene~\cite{avrahami2023break} explore unsupervised concept decomposition through localized masks within the U-Net; while effective for spatially localizable objects, such mask-centric strategies are inherently unsuitable for global attributes such as artistic style. 
In parallel, B-LoRA~\cite{frenkel2024implicit}, ZipLoRA~\cite{shah2024ziplora}, and UnZipLoRA~\cite{liu2025unziplora} tackle content--style separation more directly by jointly or independently training LoRA modules under block-wise, column-wise, and prompt-level constraints, achieving compatible content and style representations suitable for recomposition. 
Despite these advances, existing methods operate purely in the spatial or feature domain and overlook a key signal‑processing insight, namely that high frequencies carry structural edges and fine detail, while low frequencies encode global color and texture. This gap motivates our WaveletLoRA framework, which leverages multi‑scale wavelet decomposition to cleanly separate low‑frequency style from high‑frequency content.

\textbf{Frequency perspective in image generation.}
Exploiting frequency representations has benefited diverse vision tasks.
Early works incorporated the Discrete Cosine Transform (DCT) into CNNs for faster learning~\cite{ghosh2016deep} or designed frequency‑adaptive networks for super‑resolution~\cite{xie2021learning}, demonstrating the promise of spectral processing beyond spatial analysis.
In diffusion‑based generation, frequency cues have been increasingly adopted to improve synthesis quality and editing control.
For instance, FDDiff~\cite{wang2024frequency} and FDG‑Diff~\cite{fdgdiff} enhance high‑frequency components to reduce blurriness, while WaveDiff~\cite{phung2023wavelet} decomposes latent features into wavelet subbands for richer textures.
FreeU~\cite{si2024freeu} and ConsisID~\cite{yuan2025identity} modulate frequency responses to boost generation quality and identity preservation, and FCDiffusion~\cite{gao2024frequency} employs frequency‑specific pathways for multi‑modal translation.
Despite these advances, existing frequency‑based methods either rely on hand‑crafted architectural changes or auxiliary losses, and none tackle the problem of disentangling content and style from a single image. More importantly, the use of wavelet decomposition for personalized image generation remains unexplored.
WaveletLoRA fills this gap by directly conditioning LoRA branches with wavelet subbands from the reference image, thereby naturally assigning low‑frequency supervision to style and high‑frequency details to content.

\begin{figure}[t!]
    \centering
    \includegraphics[width=\textwidth]{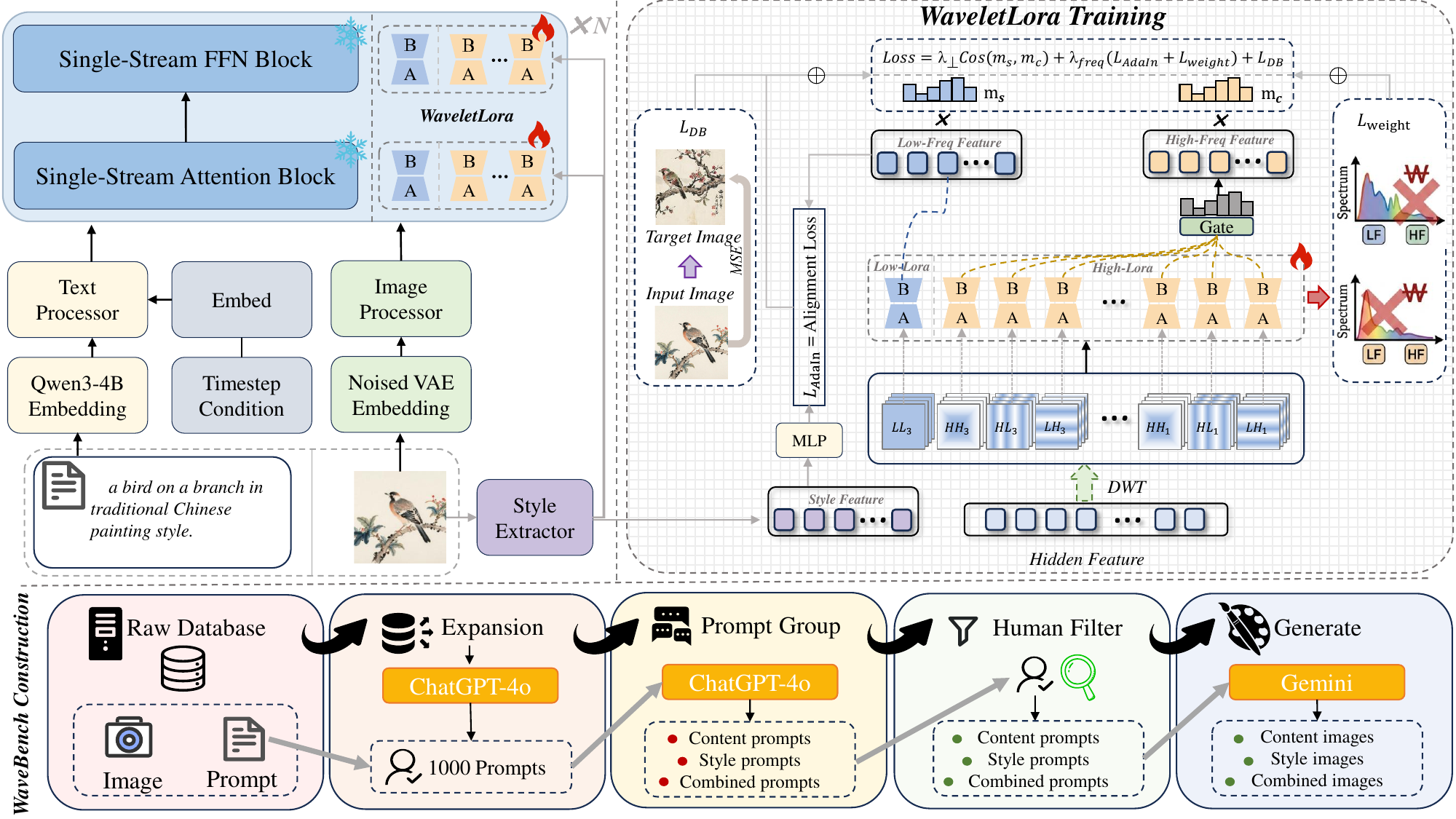}    
    \caption{\textbf{Overview of WaveletLoRA and WaveBench.} (Top) Given a reference image, a two-level DWT decomposes the image into one low-frequency subband and multiple high-frequency subbands. The low-frequency branch learns global appearance attributes for style modeling, while the high-frequency branches preserve multi-scale structural details for content representation. An adaptive aggregation MoE module fuses high-frequency representations into a unified content LoRA, followed by frequency-domain disentanglement supervision and AdaIN-based style alignment. (Bottom) Construction pipeline of WaveBench, including prompt generation, manual verification, and reference image synthesis.}
    \label{fig:waveletlora_overview}
\end{figure}
\section{Method}

\subsection{Overview of Our Method}

Given a reference image containing both subject content and visual style, our goal is to disentangle structural content from stylistic appearance for flexible personalized image generation. Existing methods typically perform disentanglement in the spatial or feature domain, often leading to residual entanglement between content and style.
To address this issue, we propose WaveletLoRA, a wavelet-guided framework for explicit content--style disentanglement. As illustrated in Fig.~\ref{fig:waveletlora_overview}, we first decompose the reference image into low-frequency and high-frequency subbands through a multi-level discrete wavelet transform (DWT). The low-frequency branch models globally coherent appearance information for style representation, while the high-frequency branches preserve structural details for content representation. The resulting frequency-specific representations are further integrated through adaptive aggregation and optimized with frequency-domain disentanglement supervision for cleaner separation between content and style.
\subsection{Wavelet-based Content–Style Decomposition}
\label{sec:wavelet_analysis}
Structural content and visual appearance often exhibit different frequency tendencies. In general, high-frequency components tend to preserve structural details such as edges and contours, whereas low-frequency components are more strongly associated with global appearance attributes including color distribution and artistic style. Motivated by this observation, we argue that frequency decomposition provides a natural inductive bias for content–style disentanglement.
% To obtain frequency-aware representations, we analyze the information encoded in each wavelet subband to motivate our frequency-guided disentanglement strategy. 
% Let $t$ be an independent variable. A mother wavelet $\psi(t)$ is scaled by $a$ and translated by $b$ to produce a family of functions $\psi_{a,b}(t)$, where the dilation parameter $a>0$ controls the width and the translation parameter $b\in\mathbb{R}$ determines the position along the $t$-axis. The continuous wavelet transform (CWT) yields a joint time-frequency representation $W(a,b)$ defined as:
% \begin{align}
% \psi_{a,b}(t) &= \frac{1}{\sqrt{a}} \psi \left( \frac{t - b}{a} \right), \quad a > 0, \quad b \in \mathbb{R}, \\
% W(a,b) &= \int_{-\infty}^{+\infty} f(t) \, \psi_{a,b}^*(t) \, dt .
% \end{align}
We employ a two-level DWT to decompose the reference image into localized frequency subbands. 
For a one-level decomposition, DWT transforms the input image into four subbands, namely $\mathbf{F} \xrightarrow{\text{DWT}} \{\mathbf{LL}_1, \mathbf{LH}_1, \mathbf{HL}_1, \mathbf{HH}_1\}$, where $\mathbf{LL}_1$ is the low-frequency approximation, and $\mathbf{LH}_1, \mathbf{HL}_1, \mathbf{HH}_1$ capture horizontal, vertical, and diagonal high-frequency details, respectively.
\begin{wrapfigure}{r}{0.5\textwidth}
    \centering
    \includegraphics[width=0.45\textwidth]{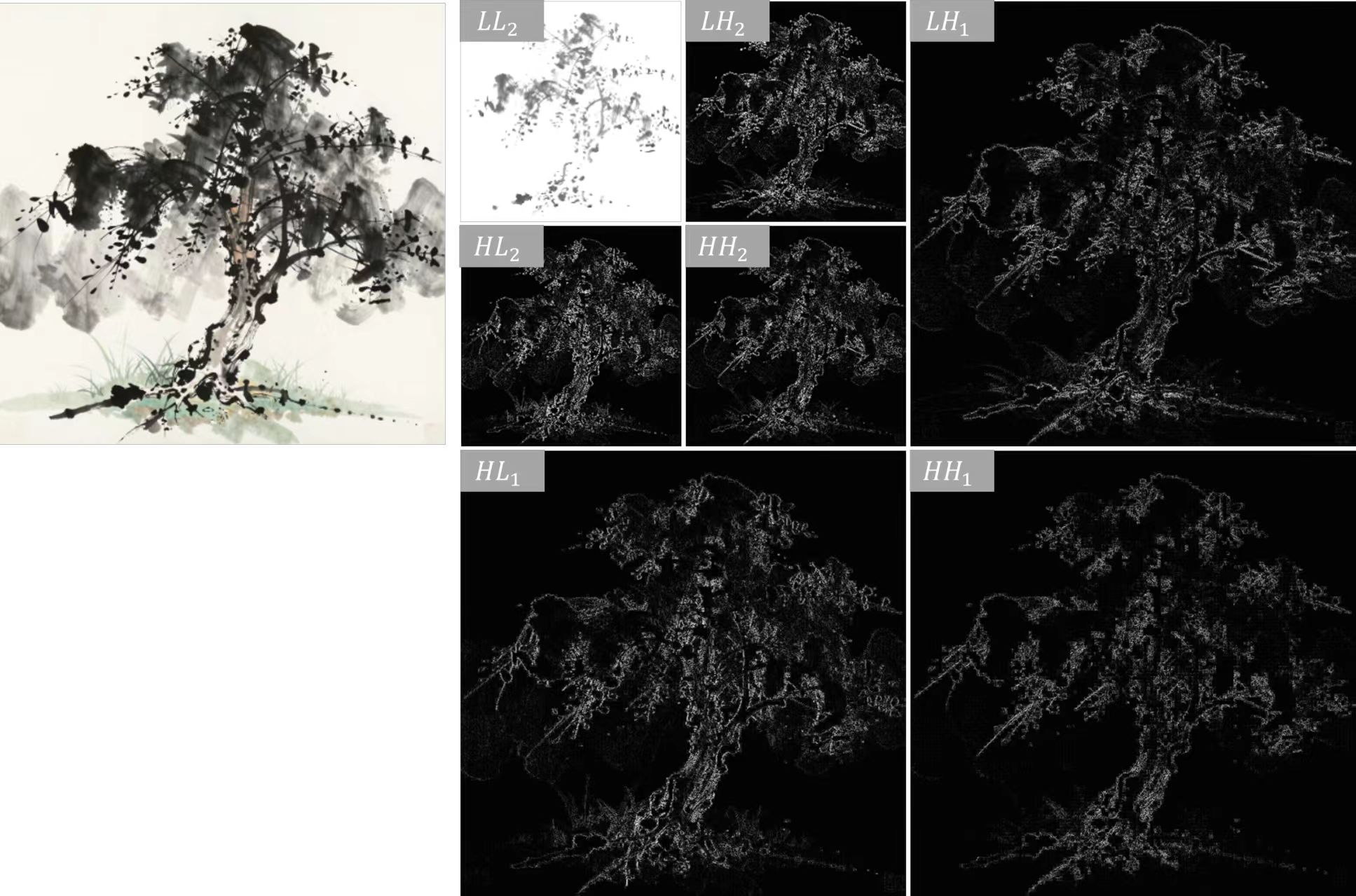}
    \caption{\textbf{Illustration of two‑level DWT.} Two-level DWT decomposes an image into one low-frequency and six high-frequency subbands at two scales. }
    \label{fig:wavelet_subbands}
\end{wrapfigure}
As illustrated in Fig.~\ref{fig:wavelet_subbands}, we further adopt a two-level DWT to decompose the image into seven subbands, the low-frequency $\mathbf{LL}_2$ and six high-frequency components ($\mathbf{LH}_{1}, \mathbf{HL}_{1}, \mathbf{HH}_{1}, \mathbf{LH}_{2}, \mathbf{HL}_{2}, \mathbf{HH}_{2}$). Among these subbands, the $\mathbf{LL}_{2}$ subband discards fine structural details but preserves global color and shading, characterizing artistic style as a supervision target for the style LoRA. In contrast, the high-frequency subbands retain multi-scale structural information, including contours, textures, and identity-related details, which are important for preserving subject structure and content consistency.
In practice, we adopt a two-level decomposition, which provides a favorable balance between global appearance modeling and structural detail preservation. As validated in Sec.~4.4, shallower decomposition retains excessive structural information in the low-frequency branch, while deeper decomposition removes informative mid-frequency structures and degrades content preservation.

\subsection{Frequency-specific LoRA Parameterization}
\label{sec:wavelet_decom}
Guided by the wavelet decomposition introduced in Sec. \ref{sec:wavelet_analysis}, we decompose the input image $I$ via a two‑level DWT into seven subbands: $\mathcal{W}(I) = \{ \text{LL}_2, \text{LH}_2, \text{HL}_2, \text{HH}_2, \text{LH}_1, \text{HL}_1, \text{HH}_1 \}$, where \(\text{LL}_2\) represents the low-frequency approximation, while the remaining six capture high-frequency details at different scales and orientations. Each subband \(S_k\) is reshaped to match the latent space dimension of the diffusion model and passed through a small learnable projector to obtain a conditioning vector \(x_k\) (of the same dimension as the text embedding).
For a diffusion backbone parameterized by weight matrix $W^{(l)}$, LoRA introduces a low-rank update $W^{(l)\prime}=W_0^{(l)}+\Delta W^{(l)}$, where $\Delta W^{(l)}=B^{(l)}A^{(l)}$, and $A^{(l)} \in \mathbb{R}^{r \times d}$, $B^{(l)} \in \mathbb{R}^{d \times r}$ are low-rank matrices with rank $r \ll d$.

Instead of learning a shared LoRA representation, we assign independent LoRA branches to different wavelet subbands. Specifically, the low-frequency branch is optimized using the $LL_2$ subband to model global appearance attributes for style representation, while the high-frequency branches are supervised by the remaining subbands to preserve structural details for content representation.
As a result, for each layer \(l\) of the DiT, we instantiate seven independent LoRA branches: $\Delta W_{\text{LL}_2}^{(l)},\; \Delta W_{\text{LH}_2}^{(l)},\; \Delta W_{\text{HL}_2}^{(l)},\; \Delta W_{\text{HH}_2}^{(l)},\; \Delta W_{\text{LH}_1}^{(l)},\; \Delta W_{\text{HL}_1}^{(l)},\; \Delta W_{\text{HH}_1}^{(l)}.$

During training, we aim for each branch to specialize in reconstructing its corresponding subband. 
For a linear layer in the DiT backbone, the adapted forward pass is
\begin{equation}
h_{\text{out}} = W_0^{(l)} h_{\text{in}} + \sum_{k=1}^{7} \Delta W_k^{(l)} \, c_k,
\end{equation} 
where \(h_{\text{in}}\) is the input hidden state and the sum runs over all \(K\) wavelet subbands, with \(c_k\) denoting the frequency‑specific conditioning vector derived from the \(k\)‑th subband.
Each update \(\Delta W_k^{(l)} = B_k^{(l)} A_k^{(l)}\) follows the standard low‑rank factorization. 
The low-frequency branch \(\Delta W_{\text{LL}_2}^{(l)}\) is designated as the \textbf{style LoRA}:
\begin{equation}
\Delta W_s^{(l)} = \Delta W_{\text{LL}_2}^{(l)}.
\end{equation}
The six high-frequency branches capture fine details and textures essential for content representation.

\subsection{Dense MoE for Adaptive Frequency Aggregation}
To integrate the six high-frequency LoRAs into a coherent content representation, we employ a dense MoE architecture with softmax gating. For each layer \(l\), we define six experts \(\{E_j^{(l)}\}_{j=1}^6\) corresponding to the high-frequency LoRAs:
\begin{equation}
E_j^{(l)} = \Delta W_{\text{HF}_j}^{(l)},\quad \text{HF}_j \in \{\text{LH}_2,\text{HL}_2,\text{HH}_2,\text{LH}_1,\text{HL}_1,\text{HH}_1\}.
\end{equation}
A gating network \(G^{(l)}(\cdot)\) computes a probability distribution over experts based on the layer input hidden states \(h\):
\begin{equation}
G^{(l)}(h) = \text{softmax}\left(W_g^{(l)} h\right),
\end{equation}
where \(W_g^{(l)}\) is a learnable matrix. The content LoRA output is then the weighted sum of expert outputs:
\begin{equation}
\Delta W_c^{(l)}(h) = \sum_{j=1}^6 G^{(l)}(h)_j \cdot E_j^{(l)}(h).
\end{equation}
The gating networks are trained jointly with all LoRA branches, allowing the model to dynamically weight different frequency experts depending on the content characteristics. This formulation enables, for example, emphasizing horizontal edges via \(\text{LH}\) experts or textures via \(\text{HH}\) experts.

% \subsubsection{Column Separation and Block Separation}
% To further enhance compatibility between the style and content LoRAs and prevent interference, we adopt two additional mechanisms. First, \textbf{column separation} introduces binary masks \(m_s\) and \(m_c\) for each LoRA column, dynamically controlling column activation during training. An importance recalibration strategy, based on gradient information, selects a sparse set of columns most relevant to each concept, ensuring that high-importance columns for style and content remain disjoint. This promotes orthogonality and reduces cross-concept contamination. Second, \textbf{block separation} leverages architectural priors by reserving specific blocks of the diffusion model predominantly for style or content learning. Inspired by observations that certain blocks exhibit higher sensitivity to either style or content, we allocate full column capacity to the corresponding LoRAs in those blocks while maintaining sparsity elsewhere. This coarse-grained separation complements the fine-grained column-level sparsity, further improving the fidelity of extracted concepts and enabling seamless recombination via direct addition.

\subsection{Training Objective}

Our full training objective combines a reconstruction loss, an orthogonality prior, and a composite frequency-domain regularization that jointly enforces weight-level frequency separation and global style-statistics alignment.

\textbf{Frequency-domain regularization loss.}
To explicitly steer the style and content branches toward low- and high-frequency specialization, we introduce a composite regularization $\mathcal{L}_{\text{freq}}$ consisting of two complementary terms.

\textit{Weight-level frequency separation.}
For each layer $l$, let $\Delta W_s^{(l)}$ and $\Delta W_c^{(l)}$ denote the style and content LoRA updates. Applying a 2D DWT to these weight matrices, we denote by $\mathcal{H}(\cdot)$ the extraction of all high-frequency subbands (LH, HL, HH) and by $\mathcal{L}(\cdot)$ the low-frequency approximation (LL). We penalize high-frequency energy in the style weights and low-frequency energy in the content weights:
\begin{equation}
\mathcal{L}_{\text{weight}} = \frac{1}{N}\sum_{l=1}^{N}\Bigl( \|\mathcal{H}(\Delta W_s^{(l)})\|_F^2 + \|\mathcal{L}(\Delta W_c^{(l)})\|_F^2 \Bigr),
\end{equation}
with $\|\cdot\|_F$ the Frobenius norm. This loss encourages the style LoRA to be smooth and the content LoRA to retain fine detail.

\textit{Style-statistics alignment.}
%While the LL$_2$ wavelet subband provides a low-frequency supervision target, we further reinforce style consistency by matching the global feature statistics of the learned style to those of the reference image. A pretrained AdaIN encoder~\cite{} extracts a style feature $\mathbf{F}_{\text{style}}$ from the input image at layer $\texttt{relu4\_1}$. Simultaneously, the LL$_2$ subband is projected through a small adapter to produce $\mathbf{F}_{\text{LL}_2}$ of the same spatial and channel dimensions. For both features, we compute channel-wise means $\boldsymbol{\mu}$ and standard deviations $\boldsymbol{\sigma}$:
To reinforce style consistency, we introduce an AdaIN~\cite{huang2017arbitrary}-based alignment mechanism. We employ a pretrained encoder to extract style features $\mathbf{F}_{\text{ref}}$ from the reference image and align them with the projected LL$2$ subband features $\mathbf{F}_{\text{LL}2}$ in the latent space.
% \begin{equation}
% \boldsymbol{\mu}(\mathbf{F})_c = \frac{1}{HW}\sum_{h,w}\mathbf{F}_{c,h,w},\qquad
% \boldsymbol{\sigma}(\mathbf{F})_c = \sqrt{\frac{1}{HW}\sum_{h,w}(\mathbf{F}_{c,h,w}-\boldsymbol{\mu}_c)^2 + \epsilon}.
% \end{equation}
The AdaIN alignment loss is then the mean squared error between the corresponding statistics:
\begin{equation}
\mathcal{L}_{\text{AdaIN}} = \frac{1}{C}\sum_{c=1}^{C}\Bigl( (\mu_{\text{LL}_2,c} - \mu_{\text{style},c})^2 + (\sigma_{\text{LL}_2,c} - \sigma_{\text{style},c})^2 \Bigr),
\end{equation}
where $C$ is the number of feature channels, and $\mu_{\text{LL}_2,c}$, $\sigma_{\text{LL}_2,c}$ (resp.\ $\mu_{\text{style},c}$, $\sigma_{\text{style},c}$) denote the per‑channel mean and standard deviation of the projected LL$_2$ features (resp.\ the style features extracted from the reference image). This term directly enforces that the global color palettes, texture granularity, and contrast of the generated style match those of the reference.

The full frequency-domain regularization is then
\begin{equation}
\mathcal{L}_{\text{freq}} = \mathcal{L}_{\text{weight}} + \lambda_{\text{AdaIN}}\,\mathcal{L}_{\text{AdaIN}}.
\end{equation}

% \textbf{Diffusion denoising loss.}
% We adopt the standard \textit{reconstruction loss} computed on the input image:
% \begin{equation}
% \mathcal{L}_{\text{DB}} = \mathbb{E}_{\boldsymbol{\epsilon}, t}\!\left[ \| \boldsymbol{\epsilon} - \boldsymbol{\epsilon}_{\theta}(\mathbf{z}_t, t, \mathbf{c}_{\text{text}}, \{\mathbf{c}_k\}) \|_2^2 \right],
% \end{equation}
% where $\mathbf{z}_t$ is a noised latent of the reference image, $t$ the diffusion timestep, $\mathbf{c}_{\text{text}}$ the text embedding, and $\{\mathbf{c}_k\}$ the set of wavelet subband embeddings injected via the LoRA branches.

\textbf{Orthogonality regularization.}
We promote disentanglement by encouraging column-wise orthogonality between the content and style LoRA masks:
\begin{equation}
\mathcal{L}_{\perp} = \sum\nolimits_l | m_c^{(l)} \cdot m_s^{(l)} |,
\end{equation}
where $m_c^{(l)}$ and $m_s^{(l)}$ are binary column masks for content and style LoRAs at layer $l$.

\textbf{Overall objective.}
The complete training loss combines a frequency-domain regularization term, a diffusion denoising loss $\mathcal{L}_{\text{DB}}$ (the standard DreamBooth reconstruction objective~\cite{ruiz2023dreambooth}), and an orthogonality regularization loss:
\begin{equation}
\mathcal{L} = \lambda_{\text{freq}} \mathcal{L}_{\text{freq}} + \lambda_{\perp} \mathcal{L}_{\perp} + \mathcal{L}_{\text{DB}},
\end{equation}
with hyperparameters $\lambda_{\text{freq}}$, $\lambda_{\text{AdaIN}}$ and $\lambda_{\perp}$ controlling the orthogonality, frequency separation, and style-statistics alignment strengths, respectively. All LoRA branches, gating networks and subband projectors are optimized jointly.

After training, the style LoRA $L_s = \{\Delta W_{\text{LL}_2}^{(l)}\}$ and the content LoRA $L_c = \{\sum_{j=1}^{6} G^{(l)}(\cdot)_j \cdot \Delta W_{\text{HF}_j}^{(l)}\}$ are extracted. They can be applied individually for style or subject transfer, or composed via direct addition $L_s \oplus L_c$ for recontextualization without further fine-tuning.

\section{Experiments}
\subsection{Experimental Settings}

% \textbf{Dataset.} We collect a diverse set of 100 stylized images covering a wide range of subjects (e.g., objects, animals, humans) and artistic styles (e.g., watercolor, cartoon, oil painting, 3D rendering). Images are sourced from online repositories, previous publications, and synthetic generations using state-of-the-art text-to-image models. Attribution details are provided in the supplementary material.

\textbf{Experimental setup.} We use Z-Image~\cite{cai2025z} as our base model. All LoRA branches are trained with rank \(r=64\) using the Adam optimizer with learning rate \(5\times10^{-5}\) for 600 steps and batch size 1. The base model weights and text encoders remain frozen. For the two-level DWT, we use the Haar wavelet. Each subband is reshaped to \(64\times64\) and projected via a small learnable MLP to match the cross-attention dimension. For column separation, we set the column cap \(N=30\%\), recalibration interval \(t=200\), and orthogonality loss weight \(\lambda_{\perp}=0.02\). The style-statistic alignment loss weight is set to \(\lambda_{\text{AdaIn}}=0.2\). During training, we use the descriptive prompts provided by WaveBench to reconstruct the reference image, whereas at test time we adopt the content‑only and style‑only prompts provided by WaveBench for decomposition evaluation. The frequency-domain loss weight is set to \(\lambda_{\text{freq}}=0.01\). All experiments are conducted on a single NVIDIA A800 GPU.

\textbf{Evaluation metrics.} Following prior work, we measure content alignment with DINOv2~\cite{hertz2022prompt} features and style alignment with CLIP-I~\cite{radford2021learning} embeddings. For each image in our WaveBench dataset (1,000 examples), we generate 4 samples per method and compute the average cosine similarity between the generated outputs and the reference image.

% \begin{figure}[h!]
%     \centering
%     \includegraphics[width=\textwidth]{figures/benchmark.png}    
%     \caption{\textbf{Diversity of WaveBench.} WaveBench comprises 1,000 diverse examples spanning a wide range of subjects (animals, objects, human figures, scenes) and styles (painting, cartoon, art, futuristic, etc.).}
%     \label{fig:benchmark}
% \end{figure}

\begin{wrapfigure}{r}{0.5\textwidth}
    \centering
    %\vspace{-3ex}
    \includegraphics[width=0.5\textwidth]{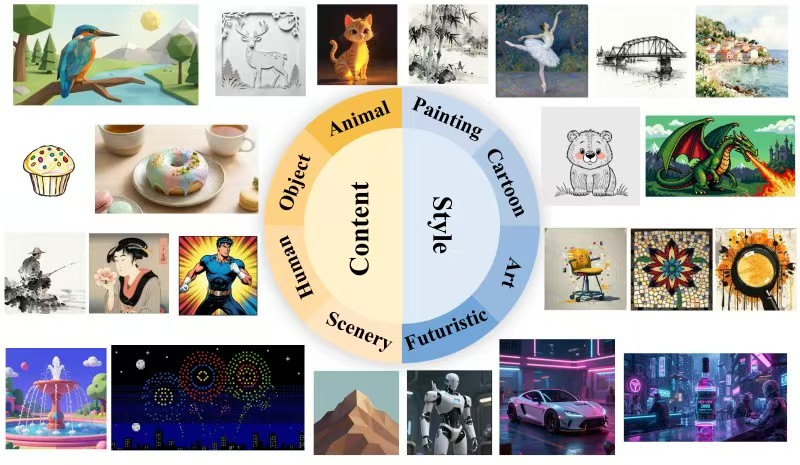}
    \caption{\textbf{Diversity of WaveBench.}}
    \label{fig:benchmark}
    %\vspace{-2ex}
\end{wrapfigure}

% \begin{table}
% \end{table}

\subsection{WaveBench} 
To rigorously evaluate single-image decomposition, we introduce WaveBench, a large-scale benchmark which comprises 1,000 diverse examples spanning a wide range of subjects (animals, objects, human figures, scenes, etc.) and styles (painting, cartoon, art, futuristic, etc.), as shown in Fig.~\ref{fig:benchmark}. Each entry includes a reference image paired with a suite of auxiliary prompts designed to test content/style transfer and fusion, offering superior scale and diversity over existing datasets. Beyond traditional feature-space metrics, we propose a VLM-based evaluation protocol using Qwen2.5-VL-72B to assess semantic disentanglement. The VLM rates two key dimensions on a $[0, 1]$ scale, namely 1) separation, which measures the removal of complementary attributes (e.g., style residue in content-only tasks), and 2) similarity, which reflects fidelity to the intended attribute. The final Content and Style Scores are computed as a weighted average of separation score and similarity score, providing a human-aligned assessment of disentanglement quality across the benchmark. Details of WaveBench and VLM-based evaluation are provided in the Appendix.

\subsection{Frequency Retention across Artistic Styles}
The frequency assignment in WaveletLoRA is a task-level inductive bias rather than a strict semantic partition. Low-frequency components tend to emphasize global appearance, while high-frequency components retain structural details. To test this tendency across diverse styles, we selected eight commonly occurring style categories from WaveBench (cartoon, Chinese painting, futuristic, oil painting, pixel art, pointillism, sketch, and water color), randomly sampled 20 images from each (160 total). For each image, we applied a two-level DWT and reconstructed an LL$_2$-only variant by zeroing the high-frequency subbands, as well as a high-only variant by zeroing LL$_2$. We then used Qwen2.5-VL-72B, a third-party VLM, to assess style and content contributions via four-choice questions. For each images, the VLM selected between one correct and three distractor options for both artistic style and subject content. Four accuracy rates were calculated: style and subject accuracy for LL$_2$-only images, and style and subject accuracy for High-only images. The results are presented in Table~\ref{tab:frequency_retention}.

\begin{table*}[ht]
\centering
\small
\caption{VLM-based recognition accuracy on frequency-retained reconstructions. LL$_2$ retains more style information, while high-frequency bands retain more subject information across all 8 categories.}
\label{tab:frequency_retention}
\begin{tabular}{lcccc}
\toprule
Style category & LL$_2$-only style & LL$_2$-only subject & High-only style & High-only subject \\
\midrule
Cartoon & 0.85 & 0.30 & 0.15 & 0.85 \\
Chinese painting & 0.80 & 0.25 & 0.10 & 0.80 \\
Futuristic & 0.90 & 0.40 & 0.25 & 0.85 \\
Oil painting & 0.70 & 0.30 & 0.05 & 0.65 \\
Pixel art & 0.85 & 0.45 & 0.20 & 0.70 \\
Pointillism & 0.75 & 0.30 & 0.05 & 0.65 \\
Sketch & 0.85 & 0.50 & 0.20 & 0.65 \\
Watercolor & 0.70 & 0.35 & 0.05 & 0.75 \\
\midrule
\tableLineColor
Average & 0.80 & 0.36 & 0.13 & 0.74 \\
\bottomrule
\end{tabular}
\end{table*}

LL$_2$-only reconstructions achieve 80\% style-recognition accuracy, compared with 13\% for high-only reconstructions. In the reverse direction, high-only reconstructions achieve 74\% subject-recognition accuracy, compared with 36\% for LL$_2$-only images. The same complementary tendency appears in each category. These results support our frequency mapping as a dominant task-level tendency.

\subsection{Experimental Results}

\begin{table}[htbp]
    \centering
    %\vspace{-2ex}
    \caption{\textbf{Quantitative comparison with state‑of‑the‑art methods.} We report feature space alignment metrics (DINOv2 for content consistency and CLIP-I for style consistency), VLM evaluation scores (content‑decom., style‑decom.), and user study scores (content, style, combined). Best results are highlighted in bold and second-best results are underlined.}
    \label{tab:method_comparison}
    %\vspace{-2ex}
    \scalebox{0.78}{
    \begin{tabular}{l *{2}{c} *{2}{c} *{3}{c}}
        \toprule
        \multicolumn{1}{c}{\multirow{2}{*}{\textbf{Method}}} & 
        \multicolumn{2}{c}{\textbf{WaveBench}} & 
        \multicolumn{2}{c}{\textbf{VLM Evaluation}} & 
        \multicolumn{3}{c}{\textbf{User Study Scores}} \\
        \cmidrule(r){2-3} \cmidrule(lr){4-5} \cmidrule(l){6-8}
        & \textbf{Content-align.}  & \textbf{Style-align.}  & \textbf{Content-decom.}  &  \textbf{Style-decom.}  & 
        \textbf{Content.}  & \textbf{Style.}  & \textbf{Combined.}  \\
        \midrule
        Qwen-Image-Edit & - & - & 0.6404 & 0.5805 & 0.5925 & 0.4025 & \textbf{0.8575} \\
        OmniGen2 & - & - & 0.5333 & 0.4252 & 0.3325 & 0.2425 & 0.7425\\
        B-LoRA & 0.5486 & 0.6115  & 0.7541 & 0.8396 & 0.5725 & 0.6250 & 0.6250\\
        UnZipLoRA & 0.5529 & \underline{0.6286} & \underline{0.8373} & 0.8693 & \underline{0.7250} &0.7800 & 0.7075\\
        USO & \textbf{0.8031} & 0.5157 & 0.6503 & \underline{0.8707} & 0.3900 & \underline{0.8325} & 0.5825\\
        \tableLineColor
        WaveletLoRA & \underline{0.5866} & \textbf{0.6330} & \textbf{0.8563} & \textbf{0.8730} & \textbf{0.7925} & \textbf{0.8400} & \underline{0.7750}\\
        \bottomrule
    \end{tabular}
    }
\end{table}

\textbf{Quantitative results.} As shown in Table~\ref{tab:method_comparison}, we compare WaveletLoRA against five representative and competitive methods, including two general-purpose instruction-tuned text-to-image models, Qwen‑Image‑Edit~\cite{wu2025qwen} and OmniGen2~\cite{wu2025omnigen2}, and three dedicated content–style decomposition approaches, B‑LoRA~\cite{frenkel2024implicit}, UnZipLoRA~\cite{liu2025unziplora}, and USO~\cite{wu2025uso}. For fair comparison, we use the same prompt templates for all methods. Specifically, content generations are guided by the content‑only prompts from WaveBench, and style generations are guided by the style‑only prompts. The two general‑purpose models are evaluated via the VLM protocol and user study, while feature‑space metrics (DINOv2 content alignment and CLIP‑I style alignment) are reported for the dedicated methods.

% \begin{wrapfigure}{r}{0.5\textwidth}
%     \centering
%     % \vspace{-3ex}
%     \includegraphics[width=0.5\textwidth]{figures/uso.jpg}
%     \vspace{-3ex}
%     \caption{\textbf{Style leakage in USO.} USO’s content‑decoupled outputs often retain strong style artifacts.}
%     \label{fig:uso_style_leakage}
%     \vspace{-2ex}
% \end{wrapfigure}

\begin{figure}[htbp]
    %\vspace{-2ex}
    \centering
    \begin{minipage}[t]{0.5\textwidth}
        \centering
        \includegraphics[width=\textwidth]{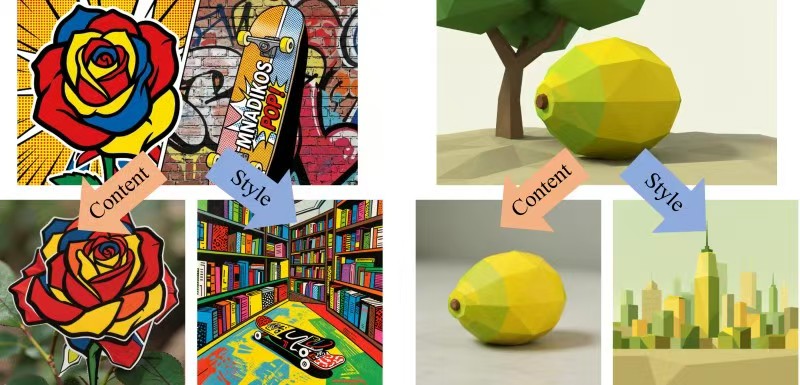}
        \caption{\textbf{Style leakage in USO.} USO's content‑decoupled outputs often retain strong style artifacts.}
        \label{fig:uso_style_leakage}
        %\vspace{-2ex}
    \end{minipage}
    \hfill
    \begin{minipage}[t]{0.45\textwidth}
        \centering
        \includegraphics[width=0.8\textwidth]{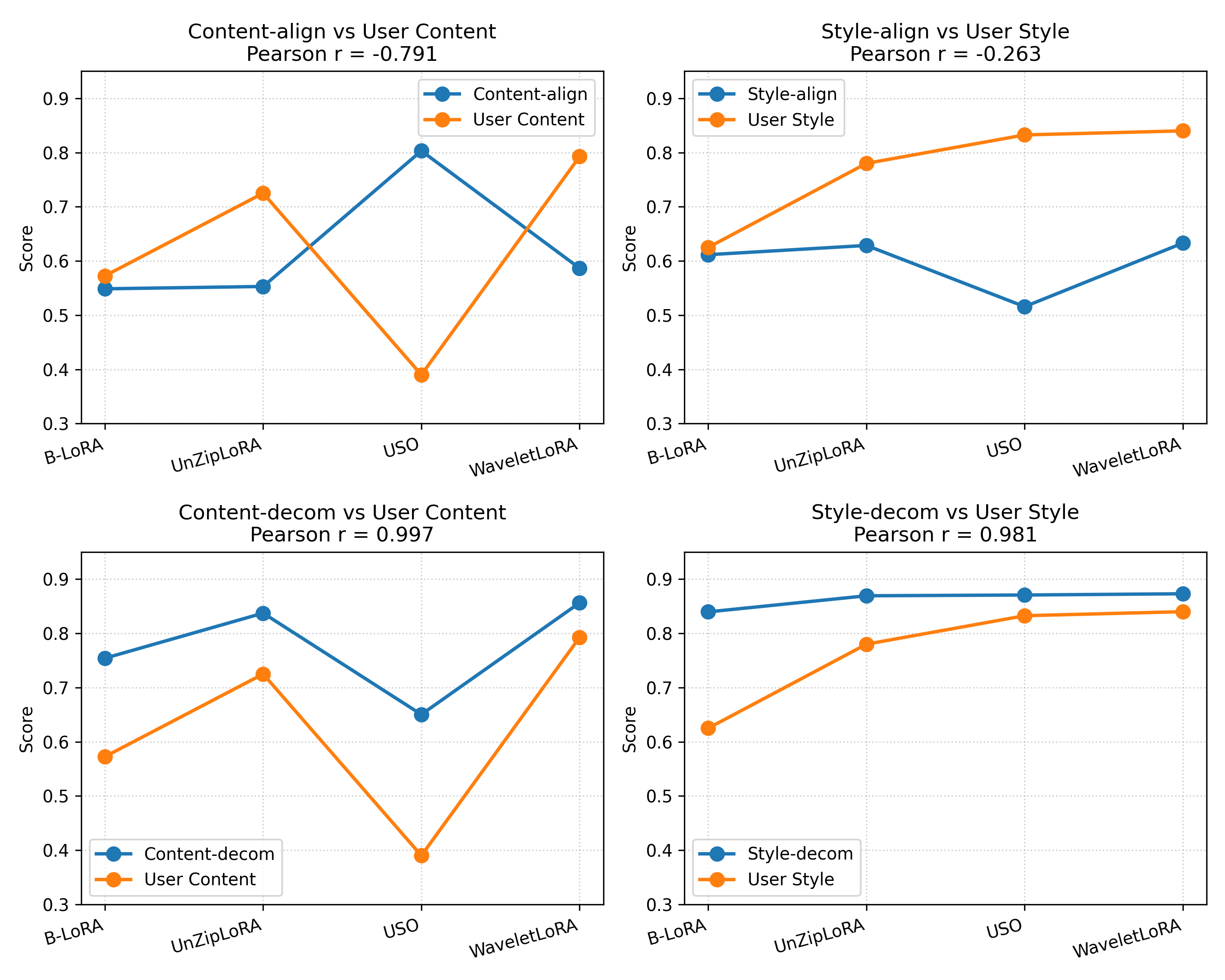}
        \caption{\textbf{Correlation between VLM content-decomposition scores and user study.}}
        \label{fig:metric_correlation}
        %\vspace{-2ex}
    \end{minipage}
\end{figure}

Table~\ref{tab:method_comparison} summarizes the quantitative results. Under feature space evaluation, USO achieves the highest DINOv2 content score ($0.8031$) but exhibits relatively weak style consistency ($0.5157$), indicating that high feature similarity does not necessarily correspond to clean disentanglement.
%USO exhibits a significant trade-off, whereas WaveletLoRA outperforms UnZipLoRA, achieving the top style score ($0.6330$). 
In contrast, WaveletLoRA achieves the best CLIP-I style consistency ($0.6330$) together with competitive content consistency ($0.5866$), demonstrating a more balanced separation between structural content and visual appearance.
Under VLM-based evaluation, WaveletLoRA achieves the best performance on both content preservation ($0.8563$) and style preservation ($0.8730$), outperforming all competing methods. This confirms that wavelet supervision enables superior disentanglement with minimal leakage. Overall, WaveletLoRA achieves the best combined result (${0.7750}$) among specialized methods, significantly surpassing USO and generic editing baselines.

We further conducted a user study to evaluate the perceptual quality of the decoupled content and style representations produced by different methods. Specifically, we randomly sampled 200 images from the test set and split them into 10 non-overlapping groups of 20 images each without replacement. 
Each group was assigned to a different participant (10 participants in total). For each image, participants were asked to rate the decoupling results of six baseline methods along three dimensions: \textit{content}, \textit{style}, and \textit{combined}. 
% All ratings were given on a 5-point Likert scale. After collection, we normalized the raw scores to the range \([0,1]\) for fair comparison across different metrics. The normalized scores are reported in Table.~\ref{tab:method_comparison}.
As shown in Table~\ref{tab:method_comparison}, all ratings utilize a 5-point Likert scale and were normalized after data collection to ensure a fair comparison.

% \begin{wrapfigure}{r}{0.5\textwidth}
%     \centering
%     \vspace{-2ex}
%     \includegraphics[width=0.5\textwidth]{figures/pearson_comparison.png}
%     \caption{\textbf{Correlation between VLM content-decomposition scores and user study.}}
%     \label{fig:metric_correlation}
%     \vspace{-2ex}
% \end{wrapfigure}

\textbf{VLM-based evaluation.} 
VLM-based evaluation aligns more closely with human judgment than conventional feature-space metrics. While DINOv2 and CLIP-I measure holistic similarity, they fail to explicitly disentangle content fidelity from style residue. 
This limitation is exemplified by USO, whose generations often retain significant stylistic traces (Fig.~\ref{fig:uso_style_leakage}). 
These residues misleadingly inflate its DINOv2 content score to a top-tier $0.8031$. In contrast, our VLM protocol directly assesses the purity of attribute separation and remains robust against such style leakage. The alignment between USO’s VLM scores and our user study further validates the reliability of this metric. 
The alignment between VLM scores and human preferences is corroborated by the correlation curves in Fig.~\ref{fig:metric_correlation}, where VLM-based metrics track human ratings more closely than DINOv2 across both content and style dimensions. These observations, supported by higher Pearson correlation coefficients, validate that our VLM protocol provides a more faithful and semantically meaningful evaluation of disentanglement than raw feature-space similarity.

\begin{wrapfigure}{r}{0.5\textwidth}
    \centering
    % \vspace{-3ex}
    \includegraphics[width=0.5\textwidth]{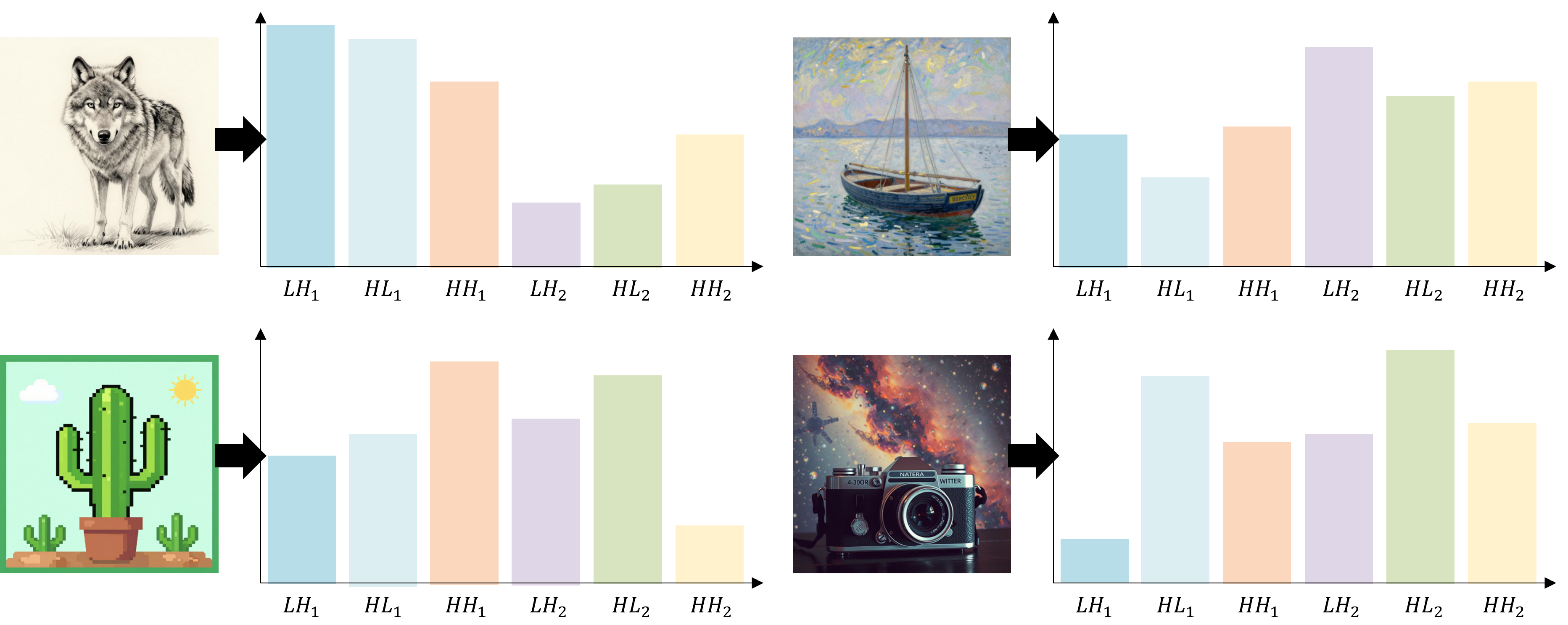}
    %\vspace{-3ex}
    \caption{\textbf{Visualization of dense MoE gating weights.} 
    % The per‑expert probabilities are shown for four images with distinct content–style combinations. The gating distribution varies substantially across samples, demonstrating that the MoE adaptively selects different high‑frequency subbands according to the input.
    }
    \label{fig:moe_gating}
\end{wrapfigure}

\textbf{Analysis of dense MoE.}
% To verify that the dense MoE module genuinely exploits frequency‑specific information rather than degenerating into a static mixture, we visualize its gating weights on four images spanning diverse content–style pairs.
% As shown in Figure. ~\ref{fig:moe_gating}, the per‑expert probability distributions differ markedly across samples.
% For the pencil‑sketch wolf, where fine hatched contours dominate, the MoE assigns higher weight to the $\text{LH}_1$ and $\text{HL}_1$ experts, which capture horizontal and vertical edge details at the original resolution.
% In the pixel‑art cactus, characterized by blocky, discrete transitions, the $\text{HH}_1$ branch receives noticeably stronger activation, consistent with its specialization in diagonal and grid‑like structures.
% The watercolor‑sketch boat and the artwork‑style camera both exhibit softer, more painterly textures; here the second‑level subbands ($\text{LH}_2$, $\text{HL}_2$, $\text{HH}_2$) are more prominently weighted, reflecting their role in encoding mid‑level shape contours and regional texture boundaries rather than pixel‑precise edges.
% This sample‑dependent variation confirms that the MoE does not collapse to a uniform average but learns to adaptively emphasize different frequency bands according to the specific content–style composition of each image.
% The resulting content LoRA selectively fuses multi‑scale structural cues without reintroducing low‑frequency style leakage, explaining the superior content‑decomposition scores observed in our quantitative evaluation.
To verify that the MoE module adaptively utilizes frequency information, we visualize its gating weights across diverse samples (Fig.~\ref{fig:moe_gating}). The distribution varies markedly by content type: for the pencil-sketch wolf, the MoE prioritizes $\text{LH}_1$ and $\text{HL}_1$ experts to capture fine-grained contours. In contrast, the pixel-art cactus triggers stronger $\text{HH}_1$ activation, targeting its blocky diagonal structures. For the watercolor boat and artwork camera, the weighting shifts to second-level subbands ($\text{LH}_2$, $\text{HL}_2$, $\text{HH}_2$) to encode mid-level textures rather than sharp edges. This sample-dependent variation confirms that our MoE avoids representation collapse, learning to selectively fuse multi-scale structural cues while preventing low-frequency style leakage.

% \textbf{User study.} We conduct a user study with 20 participants, each asked to compare our method against each baseline for both decomposition quality and recontextualization fidelity. For each comparison, participants view the input image and two sets of outputs (method A and method B, randomly ordered) consisting of four content generations and four style generations. They select which set better decomposes the input or better preserves concept fidelity in recontextualizations. As shown in Table.~\ref{tab:user_study}, WaveletLoRA is strongly preferred over all baselines, with the highest preference margins against DreamBooth-LoRA and Inspiration Tree. Notably, even compared to the strong UnZipLoRA baseline, our frequency-aware approach achieves \(71.3\%\) preference for decomposition and \(68.9\%\) for recontextualization, demonstrating the benefit of explicit frequency decomposition.

\begin{figure}[h!]
    \centering
    %\vspace{-3ex}
    \includegraphics[width=\textwidth]{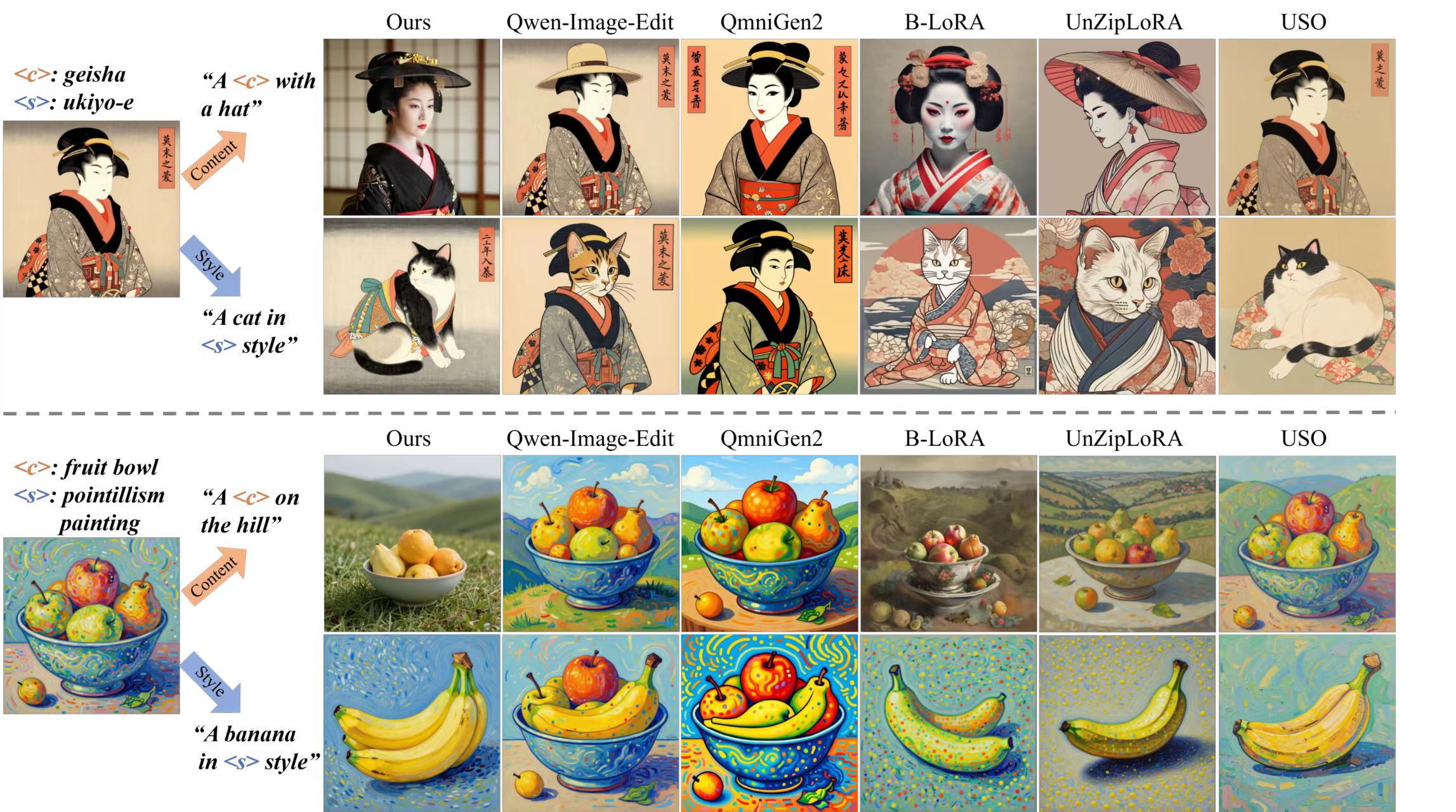}    
    \caption{\textbf{Qualitative comparison.} WaveletLoRA produces cleaner content–style separation than other methods. Content outputs (top rows) retain subject identity without style residues, and style outputs (bottom rows) capture artistic textures without subject leakage.}
    \label{fig:qualitative_comparison}
    %\vspace{-2ex}
\end{figure}

% \subsection{Qualitative Results}

% As shown in Fig.~\ref{fig:qualitative_comparison}, existing approaches often leave detectable style residues in content-decoupled outputs or fail to fully preserve subject identity when style is removed. 
% For instance, with a watercolor-style portrait, B-LoRA retains visible watercolor washes in the content-only generation, blending the subject with the original medium, while UnZipLoRA erodes fine facial contours and USO introduces noticeable color artifacts inherited from the source style. 
% In contrast, our method reconstructs the subject with sharp structural fidelity and a neutral appearance, successfully discarding the watercolor texture through explicit low-frequency style separation. 
% These examples demonstrate that by supervising style and content branches with distinct wavelet subbands and enforcing frequency-level separation, WaveletLoRA effectively isolates structure from appearance, producing more faithful and reusable representations.
% Similarly, for a 3D-rendered animal, the style outputs of baseline methods typically carry ghosting of the original animal silhouette or mismatched textures, whereas WaveletLoRA extracts a cohesive 3D texture and lighting pattern that can be applied to entirely different subjects without shape leakage. 

\textbf{Qualitative results.} As shown in Fig.~\ref{fig:qualitative_comparison}, existing methods often leave style residues in content outputs or lose subject identity. For example, in the first row, UnZipLoRA, USO and other methods still retain obvious watercolor textures and stylized color patterns in the generated woman image, leading to unrealistic facial appearance and blurred structural details. In contrast, WaveletLoRA reconstructs a more realistic face with cleaner contours and more faithful identity-related structures while effectively removing the original artistic style. Similar observations can also be found in the second row. Moreover, the style generation results further demonstrate the effectiveness of our frequency-aware disentanglement. These results demonstrate that wavelet-domain supervision enables cleaner disentanglement between structural content and visual style.

\subsection{Ablation Study}

We evaluate the contribution of each component by assessing content and style alignment on WaveBench, utilizing DINOv2 and CLIP-I respectively.

\textbf{Decomposition depth.}
Replacing the two‑level DWT with a single‑level transform raises content alignment to $0.5904$ but severely degrades style alignment to $0.6112$. The drop stems from residual structural contours in the LL$_1$ subband, which contaminate the style LoRA and produce an inflated content score through style leakage. A three‑level DWT excessively filters mid‑frequency details, lowering content alignment to $0.5788$ without style gains. Our two‑level design thus strikes the optimal balance.

\begin{wraptable}{r}{0.45\textwidth}
    \centering
    \small
    %\vspace{-10pt} % 缩小上方间距
    \caption{\textbf{Ablation on decomposition depth and fusion strategy.}}
    \label{tab:ablation}
    \scalebox{0.8}{
    \begin{tabular}{lcc}
        \toprule
        \textbf{Configuration} & {\textbf{Content-align.}} & {\textbf{Style-align.}} \\
        \midrule
        \tableLineColor \textbf{WaveletLoRA} & \textbf{0.5866} & \textbf{0.6330} \\
        \midrule
        \textit{Decomposition Depth} \\
        \quad 1-level DWT & 0.5904 & 0.6112 \\
        \quad 3-level DWT & 0.5788 & 0.6317 \\
        \noalign{\smallskip}\hdashline[1pt/2pt]\noalign{\smallskip}
        \textit{High-frequency Fusion} \\
        \quad Equal averaging & 0.5752 & 0.6329 \\
        \quad Learnable scalars $\alpha_j$ & 0.5769 & 0.6324 \\
        \quad Cross-attention & 0.5831 & 0.6326 \\
        \noalign{\smallskip}\hdashline[1pt/2pt]\noalign{\smallskip}
        \textit{Style Feature Extractor} \\
        \quad w/o AdaIN & 0.5863 & 0.6321 \\
        \bottomrule
    \end{tabular}
    }
    \vspace{4ex}
    \caption{\textbf{AdaIN ablation on the pointillism subset of WaveBench.}}
    \label{tab:pointillism_adain}
    \begin{tabular}{lcc}
        \toprule
        Full model & w/o AdaIN & Drop \\
        \midrule
        0.703 & 0.655 & 0.048 \\
        \bottomrule
    \end{tabular}
\end{wraptable}

\textbf{High‑frequency fusion.}
Equal averaging causes the largest content drop ($0.5752$), as it ignores spatial and temporal relevance. Learnable per‑band scalars improve content to $0.5769$, but remain globally fixed. Cross‑attention reaches $0.5831$ by dynamically conditioning on layer features, yet it incurs extra projections and lacks expert‑specialised gating. Our dense MoE achieves the best content alignment ($0.5866$) while preserving style, confirming the advantage of input‑dependent, expert‑wise weighting.

\textbf{Style regularization.}
Removing the AdaIN loss marginally reduces style alignment ($0.6321$), validating its role as a lightweight style refinement. In particular, for styles whose defining characteristics may be concentrated in high-frequency components, our AdaIN-based style alignment provides complementary supervision for global style statistics that may not be fully captured by the LL$_2$-based objective. To validate this, we conducted an additional ablation on the pointillism subset of WaveBench using Qwen2.5-VL-72B style scores. As shown in Table~\ref{tab:pointillism_adain}, removing AdaIN lowers the style score from 0.703 to 0.655, a drop of 0.048. This result confirms that AdaIN contributes meaningfully to style fidelity on this highly textured subset.

Moreover, we further analyze the sensitivity of the framework to the two key loss weights, $\lambda_{\mathrm{freq}}$ and $\lambda_{\mathrm{AdaIN}}$, as reported in appendix.

% \subsection{Applications and Extensions}

% \textbf{Cross-image composition.} WaveletLoRA enables cross-composition of style and content from different images. Given style LoRA from image \(A\) and content LoRA from image \(B\), we can directly add them (without retraining) to generate novel combinations. Figure.~\ref{fig:cross} shows examples where the low-frequency style of one image is applied to the high-frequency content of another, yielding plausible and aesthetically pleasing results. This capability arises from the orthogonal frequency decomposition enforced during training.

% \textbf{Failure cases.} WaveletLoRA may struggle when the input image contains highly abstract styles that severely distort subject shapes (e.g., cubism), leading to residual style patterns in the content LoRA. Addressing such cases through adaptive wavelet bases or multi-level decomposition remains future work.

\section{Conclusion}
% We presented WaveletLoRA, a frequency‑aware framework that disentangles content and style from a single image by supervising dedicated LoRA branches with multi‑scale wavelet subbands. A low‑frequency branch captures style, while high‑frequency branches fused via a dense MoE preserve content, further purified by a frequency disentanglement loss and an AdaIN style‑statistics term. Extensive experiments on WaveBench show that WaveletLoRA outperforms existing methods in both feature‑space and VLM‑based evaluation, and the decomposed LoRAs enable seamless cross‑image composition.
We present WaveletLoRA, a frequency-aware framework for content--style disentanglement in personalized image generation. By introducing wavelet-domain supervision, our method explicitly separates low-frequency appearance information from high-frequency structural details through frequency-specific LoRA branches. We further establish WaveBench and a VLM-based evaluation protocol for disentanglement assessment. Extensive experiments demonstrate that WaveletLoRA achieves cleaner content-style separation and more realistic generation results than existing methods.

\begin{ack}
This research is supported by Artificial Intelligence-National Science and Technology Major Project (2023ZD0121200), the National Natural Science Foundation of China (62531026, 62437001, 62436001), the Natural Science Foundation of Jiangsu Province under Grant BK20243051, and the Strategic Priority Research Program of Chinese Academy of Sciences under Grant XDB1350103.
\end{ack}

\bibliographystyle{plain}
\bibliography{main}

\appendix

\section{WaveBench} 
\paragraph{WaveBench Construction.}To rigorously benchmark single-image content--style decomposition, we construct WaveBench, a large-scale dataset of 1,000 diverse and carefully curated examples. As illustrated in Fig.~\ref{fig:waveletlora_overview}, the construction follows a multi-stage pipeline:
\begin{itemize}
\item \textbf{Raw Database \& Expansion:} The initial prompt pool is aggregated from CSD-100~\cite{nguyen2025csd} (100 prompts) and 3SGen~\cite{song20253sgen} (135 prompts). To ensure scale and diversity, we employ \textbf{ChatGPT-4o} to expand the collection with 765 additional unique prompts, reaching a total of 1,000 candidate entries.
\item \textbf{Prompt Grouping:} For each original prompt $P_i^{\text{orig}}$, ChatGPT-4o acts as a semantic engine to generate a structured prompt group. This includes: (1) two \textbf{content-only} prompts $\{P_{i,j}^{\text{cont}}\}_{j=1}^{2}$ that port the original subject into new contexts; (2) two \textbf{style-only} prompts $\{P_{i,k}^{\text{sty}}\}_{k=1}^{2}$ that apply the style to disparate subjects; and (3) two \textbf{combined} prompts $\{P_{i,\ell}^{\text{comb}}\}_{\ell=1}^{2}$ for holistic reconstruction.
\item \textbf{Human Filter:} Every prompt group undergoes rigorous manual inspection. Human annotators verify semantic correctness, eliminate duplicates, and ensure a broad coverage of artistic styles and complex subjects, reducing potential evaluation noise.
\item \textbf{Reference Generation:} Finally, the verified $P_i^{\text{orig}}$ is fed into \textbf{Gemini-3 Pro}~\cite{google2025gemini3pro} to synthesize high-fidelity reference images $I_i$ ($1024 \times 1024$).
\end{itemize}

Each WaveBench entry is formally defined as an \textbf{8-tuple}:
\begin{equation}
\mathcal{D}_i =
\bigl(
P_i^{\text{orig}}, I_i,
\{P_{i,j}^{\text{cont}}\}_{j=1}^{2},
\{P_{i,k}^{\text{sty}}\}_{k=1}^{2},
\{P_{i,\ell}^{\text{comb}}\}_{\ell=1}^{2}
\bigr).
\end{equation}

WaveBench significantly advances previous content-style decomposition benchmarks in scale, annotation quality, and diversity. As visualized in Figure.~\ref{fig:benchmark}, with 1,000 curated examples, it is significantly larger than existing datasets, providing a more reliable foundation for statistical comparison.
Manual screening at every stage guarantees clean, unambiguous annotations, reducing evaluation noise and providing a faithful testbed for disentanglement methods.

\paragraph{WaveBench Dataset Statistics}
WaveBench contains 1,000 examples. Table~\ref{tab:wavebench_distribution} summarizes the top-level style and content labels.

\begin{table*}[ht]
\centering
\small
\caption{Top-level style and content category distributions in WaveBench.}
\label{tab:wavebench_distribution}
\begin{minipage}[t]{0.45\textwidth}
\centering
\begin{tabular}{lc}
\toprule
Style category & Count \\
\midrule
Painting & 426 \\
Cartoon & 97 \\
Futuristic & 171 \\
Art & 128 \\
Other & 178 \\
\midrule
Total & 1,000 \\
\bottomrule
\end{tabular}
\end{minipage}\hfill
\begin{minipage}[t]{0.45\textwidth}
\centering
\begin{tabular}{lc}
\toprule
Content category & Count \\
\midrule
Animals & 195 \\
Human figures & 157 \\
Landscapes & 202 \\
Objects & 222 \\
Other & 224 \\
\midrule
Total & 1,000 \\
\bottomrule
\end{tabular}
\end{minipage}
\end{table*}

\paragraph{VLM‑based evaluation.}
Leveraging WaveBench, we design an automatic evaluation protocol powered by Qwen2.5‑VL‑72B~\cite{bai2025qwen2.5}, a large vision-language model, to assess disentanglement in a human‑aligned, semantic manner.

For a given method and each sample $\mathcal{D}_i$, we generate two content‑decoupled images: one for each $P_{i,1}^{\text{cont}}$ and $P_{i,2}^{\text{cont}}$, denoted $\{I_{i,1}^{\text{cont}}, I_{i,2}^{\text{cont}}\}$. Similarly, we obtain two style‑decoupled images $\{I_{i,1}^{\text{sty}}, I_{i,2}^{\text{sty}}\}$ from the two style‑only prompts.
The VLM is queried with the reference $I_i$ paired with each generated image to rate two aspects on a $[0,1]$ scale:
\begin{itemize}
  \item \textbf{Separation}: how well the complementary attribute has been removed (style residue for a content‑decoupled image, subject residue for a style‑decoupled image);
  \item \textbf{Similarity}: how similar the two images are in terms of the intended attribute.
\end{itemize}
For a content‑decoupled image $I^{\text{cont}}$, let $s_{\text{sep}}(I_i, I^{\text{cont}})$ be the style‑separation score and $s_{\text{subj}}(I_i, I^{\text{cont}})$ the subject‑similarity score. The per‑image content score is
\begin{equation}
  s_{\text{cont}}(I_i, I^{\text{cont}}) = 0.6 \cdot s_{\text{sep}}(I_i, I^{\text{cont}}) + 0.4 \cdot s_{\text{subj}}(I_i, I^{\text{cont}}).
\end{equation}
The sample‑level content score is the average over the two generated content images:
\begin{equation}
  s_{\text{cont}}(\mathcal{D}_i) = \frac{1}{2}\sum_{m=1}^{2} s_{\text{cont}}(I_i, I_{i,m}^{\text{cont}}).
\end{equation}
Analogously, for a style‑decoupled image $I^{\text{sty}}$ we obtain a subject‑separation score $s_{\text{sep}}'$ and a style‑similarity score $s_{\text{sty}}'$. We define the per‑image style score as
\begin{equation}
  s_{\text{sty}}(I_i, I^{\text{sty}}) = 0.6 \cdot s_{\text{sep}}'(I_i, I^{\text{sty}}) + 0.4 \cdot s_{\text{sty}}'(I_i, I^{\text{sty}}),
\end{equation}
and average across the two style generations:
\begin{equation}
  s_{\text{sty}}(\mathcal{D}_i) = \frac{1}{2}\sum_{n=1}^{2} s_{\text{sty}}(I_i, I_{i,n}^{\text{sty}}).
\end{equation}
The final \textbf{Content Score} and \textbf{Style Score} reported for each method are the averages of $s_{\text{cont}}(\mathcal{D}_i)$ and $s_{\text{sty}}(\mathcal{D}_i)$ over all $1{,}000$ WaveBench samples, respectively.
This VLM‑based metric directly evaluates semantic disentanglement, complementing traditional feature‑space measures.

\section{Additional Evaluation Analyses}

\paragraph{Cross-VLM Evaluation}
To test sensitivity to the choice of evaluator, we evaluated all methods on the same samples with the same protocol using Gemini-3.5-Flash and GPT-5.1. WaveletLoRA obtains the highest score on both dimensions under all three VLMs (Table~\ref{tab:cross_vlm}), indicating that the main comparison is not specific to one evaluator.

\begin{table*}[ht]
\centering
\small
\caption{Evaluation scores from three VLMs under the same protocol.}
\label{tab:cross_vlm}
\resizebox{\textwidth}{!}{%
\begin{tabular}{lcccccc}
\toprule
& \multicolumn{2}{c}{Qwen2.5-VL-72B} & \multicolumn{2}{c}{Gemini-3.5-Flash} & \multicolumn{2}{c}{GPT-5.1} \\
Method & Content & Style & Content & Style & Content & Style \\
\midrule
Qwen-Image-Edit & 0.6404 & 0.5805 & 0.6103 & 0.5724 & 0.6152 & 0.5920 \\
OmniGen2 & 0.5333 & 0.4252 & 0.5169 & 0.3985 & 0.5087 & 0.4003 \\
B-LoRA & 0.7541 & 0.8396 & 0.7346 & 0.8450 & 0.7326 & 0.8433 \\
UnZipLoRA & 0.8373 & 0.8693 & 0.8113 & 0.8584 & 0.7945 & 0.8608 \\
USO & 0.6503 & 0.8707 & 0.5752 & 0.8601 & 0.6032 & 0.8607 \\
\tableLineColor
WaveletLoRA & 0.8563 & 0.8730 & 0.8379 & 0.8667 & 0.7995 & 0.8642 \\
\bottomrule
\end{tabular}%
}
\end{table*}

\paragraph{Training and Inference Cost}
Table~\ref{tab:compute_cost} reports the measured training time for 600 optimization steps, inference time per generated image, and additional trainable parameters. WaveletLoRA incurs additional computation from its frequency-specific branches and AdaIN component.

\begin{table*}[ht]
\centering
\small
\caption{Computational cost comparison.}
\label{tab:compute_cost}
\begin{tabular}{lccc}
\toprule
Component & Training time & Inference time & Additional parameters \\
\midrule
Base Z-Image (without LoRA) & -- & $\sim$5 s & 0 \\
Base + style and content LoRA & $\sim$230 min & $\sim$6 s & $2\times(2\times64\times d)$ \\
WaveletLoRA (full) & $\sim$320 min & $\sim$10 s & $7\times(2\times64\times d)+6\times d+\mathrm{AdaIN}$ \\
\bottomrule
\end{tabular}
\end{table*}

\paragraph{Parameter Sensitivity Analysis}
\label{parameter_senstivity}
\begin{table}[htbp]
\centering
\begin{minipage}[t]{0.48\linewidth}
\centering
\caption{Sensitivity Analysis of $\lambda_{freq}$}
\label{tab:sens_freq}
\vspace{0.5em}
\begin{tabular}{l|cc}
\toprule
$\lambda_{freq}$ & Content-align. & Style-align. \\
\midrule
0.002 & 0.5860 & 0.6298 \\
0.006 & 0.5863 & 0.6304 \\
\tableLineColor 0.01 & 0.5866 & 0.6330 \\
0.02 & 0.5862 & 0.6324 \\
0.04 & 0.5862 & 0.6317 \\
\bottomrule
\end{tabular}
\end{minipage}
\hfill
\begin{minipage}[t]{0.48\linewidth}
\centering
\caption{Sensitivity Analysis of $\lambda_{AdaIn}$}
\label{tab:sens_ada}
\vspace{0.5em}
\begin{tabular}{l|cc}
\toprule
$\lambda_{AdaIn}$ & Content-align. & Style-align. \\
\midrule
0.05 & 0.5865 & 0.6325 \\
0.1 & 0.5866 & 0.6327 \\
\tableLineColor 0.2 & 0.5866 & 0.6330 \\
0.3 & 0.5865 & 0.6320 \\
0.4 & 0.5867 & 0.6326 \\
0.5 & 0.5866 & 0.6322 \\
0.6 & 0.5865 & 0.6317 \\
\bottomrule
\end{tabular}
\end{minipage}
\end{table}

We analyze the sensitivity of the framework to the two key loss weights, $\lambda_{\mathrm{freq}}$ and $\lambda_{\mathrm{AdaIN}}$, as reported in Table~\ref{tab:sens_freq} and Table~\ref{tab:sens_ada}.
For the frequency disentanglement loss, $\lambda_{\mathrm{freq}}=0.01$ attains the highest content alignment ($0.5866$) and the peak style alignment ($0.6330$).
Smaller weights weaken the regularization that penalizes high‑frequency energy in the style LoRA, whereas larger weights overly suppress the style branch's capacity, degrading content alignment without further style gains.
For the AdaIN loss, $\lambda_{\mathrm{AdaIN}}=0.2$ yields the optimal style alignment ($0.6330$) while preserving content alignment ($0.5866$).
Excessively large $\lambda_{\mathrm{AdaIN}}$ (e.g., $0.6$) forces the style branch to mimic the reference statistics too aggressively, leading to a decline in style alignment ($0.6317$) and no content advantage.
Consequently, we adopt $\lambda_{\mathrm{freq}}=0.01$ and $\lambda_{\mathrm{AdaIN}}=0.2$ throughout our experiments, which strikes a favorable balance between content preservation and style fidelity. 

\section{More Qualitative Results}
\begin{figure}[h!]
    \centering
    \includegraphics[width=\textwidth]{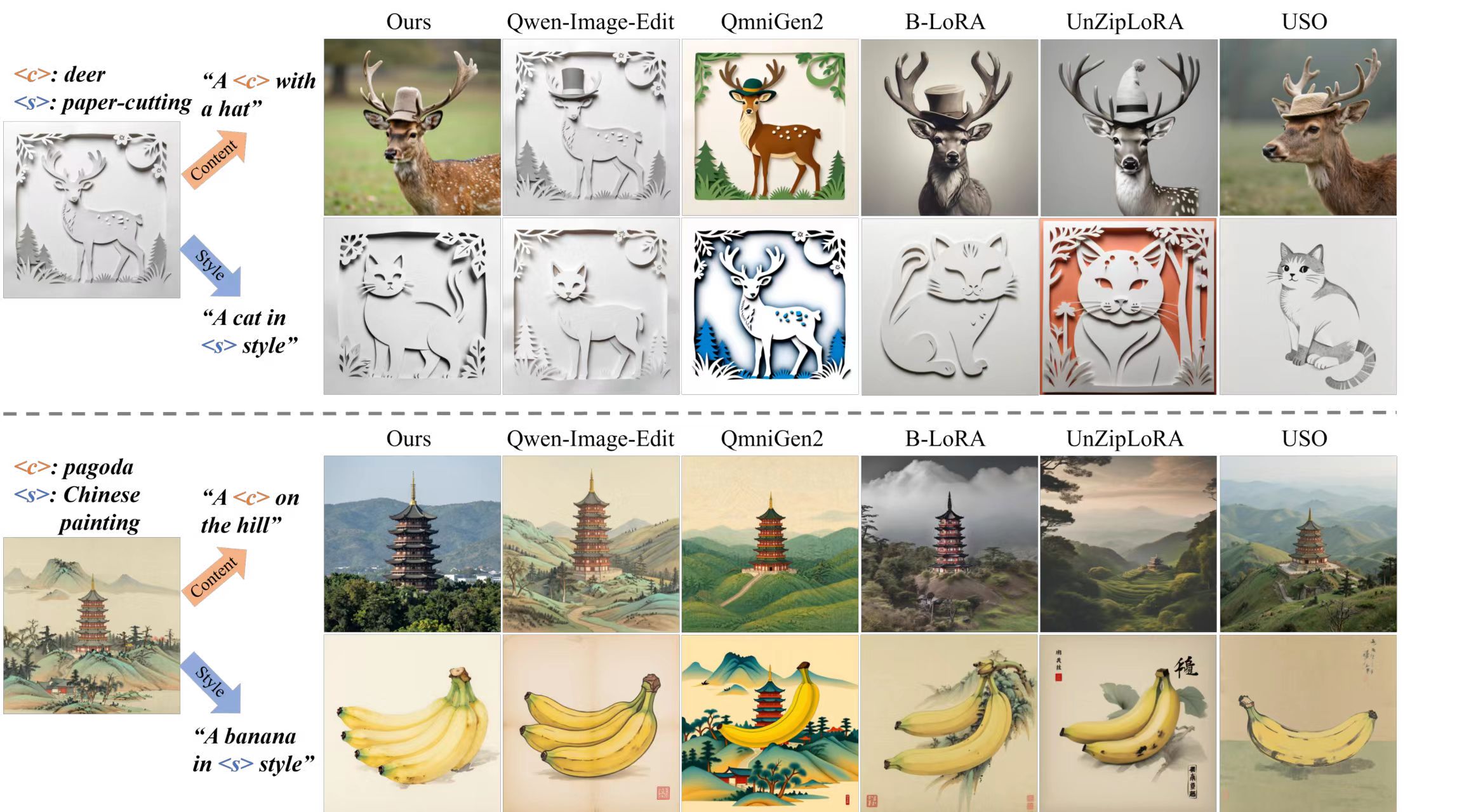}    
    \caption{\textbf{Additional qualitative comparisons across diverse content–style pairs.} We show decomposition results on other set of subjects and artistic styles. For each example, the top row displays content‑decoupled outputs and the bottom row shows style‑decoupled outputs. WaveletLoRA consistently preserves subject identity without ghosting from the original style and extracts coherent artistic textures without silhouette leakage from the source content.}
    \label{fig:qualitative_appendix}
\end{figure}

As shown in Figure~\ref{fig:qualitative_appendix}, we extend our qualitative evaluation to a broader suite of content–style combinations beyond those presented in the main text. Across this spectrum, WaveletLoRA exhibits consistent and robust decomposition behavior. Content‑decoupled generations faithfully retain the original subject's identity, pose, and fine structural details, while successfully discarding the color palettes, textures, and brushwork of the reference style. Style‑decoupled outputs capture the defining artistic characteristics—stroke patterns, shading conventions, and palette distributions—and transfer them cleanly to new textual prompts without carrying over silhouettes or object boundaries from the source image. Baselines such as USO and UnZipLoRA, by contrast, show occasional style bleeding in content outputs or subject residuals in style outputs, particularly under strong stylization (e.g., heavy brush textures or low‑contrast line drawings). These qualitative observations align with the quantitative metrics reported in the main paper and further confirm that the frequency‑supervised design of WaveletLoRA enables more thorough and generalizable content–style separation across a wide range of visual domains.

%%%%%%%%%%%%%%%%%%%%%%%%%%%%%%%%%%%%%%%%%%%%%%%%%%%%%%%%%%%%

\newpage
\section*{NeurIPS Paper Checklist}

\begin{enumerate}

\item {\bf Claims}
    \item[] Question: Do the main claims made in the abstract and introduction accurately reflect the paper's contributions and scope?
    \item[] Answer: \answerYes{} % Replace by \answerYes{}, \answerNo{}, or \answerNA{}.
    \item[] Justification: See Abstract and Section~1.

\item {\bf Limitations}
    \item[] Question: Does the paper discuss the limitations of the work performed by the authors?
    \item[] Answer: \answerYes{} % Replace by \answerYes{}, \answerNo{}, or \answerNA{}.
    \item[] Justification: See Section~4.4.

\item {\bf Theory assumptions and proofs}
    \item[] Question: For each theoretical result, does the paper provide the full set of assumptions and a complete (and correct) proof?
    \item[] Answer: \answerYes{} % Replace by \answerYes{}, \answerNo{}, or \answerNA{}.
    \item[] Justification: See Section~3.

    \item {\bf Experimental result reproducibility}
    \item[] Question: Does the paper fully disclose all the information needed to reproduce the main experimental results of the paper to the extent that it affects the main claims and/or conclusions of the paper (regardless of whether the code and data are provided or not)?
    \item[] Answer: \answerYes{} % Replace by \answerYes{}, \answerNo{}, or \answerNA{}.
    \item[] Justification: See Section~4 and Appendix.

\item {\bf Open access to data and code}
    \item[] Question: Does the paper provide open access to the data and code, with sufficient instructions to faithfully reproduce the main experimental results, as described in supplemental material?
    \item[] Answer: \answerYes{} % Replace by \answerYes{}, \answerNo{}, or \answerNA{}.
    \item[] Justification: See Supplementary Material.

\item {\bf Experimental setting/details}
    \item[] Question: Does the paper specify all the training and test details (e.g., data splits, hyperparameters, how they were chosen, type of optimizer) necessary to understand the results?
    \item[] Answer: \answerYes{} % Replace by \answerYes{}, \answerNo{}, or \answerNA{}.
    \item[] Justification: See Section~4 and Appendix.

\item {\bf Experiment statistical significance}
    \item[] Question: Does the paper report error bars suitably and correctly defined or other appropriate information about the statistical significance of the experiments?
    \item[] Answer: \answerYes{} % Replace by \answerYes{}, \answerNo{}, or \answerNA{}.
    \item[] Justification: See Section~4.

\item {\bf Experiments compute resources}
    \item[] Question: For each experiment, does the paper provide sufficient information on the computer resources (type of compute workers, memory, time of execution) needed to reproduce the experiments?
    \item[] Answer: \answerYes{} % Replace by \answerYes{}, \answerNo{}, or \answerNA{}.
    \item[] Justification: See Section~4.
    
\item {\bf Code of ethics}
    \item[] Question: Does the research conducted in the paper conform, in every respect, with the NeurIPS Code of Ethics \url{https://neurips.cc/public/EthicsGuidelines}?
    \item[] Answer: \answerYes{} % Replace by \answerYes{}, \answerNo{}, or \answerNA{}.
    \item[] Justification: See all sections of the paper.

\item {\bf Broader impacts}
    \item[] Question: Does the paper discuss both potential positive societal impacts and negative societal impacts of the work performed?
    \item[] Answer: \answerYes{} % Replace by \answerYes{}, \answerNo{}, or \answerNA{}.
    \item[] Justification: See Section~1 and Section~2.
    
\item {\bf Safeguards}
    \item[] Question: Does the paper describe safeguards that have been put in place for responsible release of data or models that have a high risk for misuse (e.g., pre-trained language models, image generators, or scraped datasets)?
    \item[] Answer: \answerNA{} % Replace by \answerYes{}, \answerNo{}, or \answerNA{}.
    \item[] Justification: Our paper poses no such risks.

\item {\bf Licenses for existing assets}
    \item[] Question: Are the creators or original owners of assets (e.g., code, data, models), used in the paper, properly credited and are the license and terms of use explicitly mentioned and properly respected?
    \item[] Answer: \answerYes{} % Replace by \answerYes{}, \answerNo{}, or \answerNA{}.
    \item[] Justification: See all sections of the paper.

\item {\bf New assets}
    \item[] Question: Are new assets introduced in the paper well documented and is the documentation provided alongside the assets?
    \item[] Answer: \answerYes{} % Replace by \answerYes{}, \answerNo{}, or \answerNA{}.
    \item[] Justification: See Section~4 and Supplementary Material.

\item {\bf Crowdsourcing and research with human subjects}
    \item[] Question: For crowdsourcing experiments and research with human subjects, does the paper include the full text of instructions given to participants and screenshots, if applicable, as well as details about compensation (if any)? 
    \item[] Answer: \answerYes{} % Replace by \answerYes{}, \answerNo{}, or \answerNA{}.
    \item[] Justification: See Appendix.

\item {\bf Institutional review board (IRB) approvals or equivalent for research with human subjects}
    \item[] Question: Does the paper describe potential risks incurred by study participants, whether such risks were disclosed to the subjects, and whether Institutional Review Board (IRB) approvals (or an equivalent approval/review based on the requirements of your country or institution) were obtained?
    \item[] Answer: \answerNA{} % Replace by \answerYes{}, \answerNo{}, or \answerNA{}.
    \item[] Justification: Our paper does not involve crowdsourcing nor research with human subjects.

\item {\bf Declaration of LLM usage}
    \item[] Question: Does the paper describe the usage of LLMs if it is an important, original, or non-standard component of the core methods in this research? Note that if the LLM is used only for writing, editing, or formatting purposes and does \emph{not} impact the core methodology, scientific rigor, or originality of the research, declaration is not required.
    %this research? 
    \item[] Answer: \answerNA{} % Replace by \answerYes{}, \answerNo{}, or \answerNA{}.
    \item[] Justification: The core method development in this research does not involve LLMs as any important, original, or non-standard components.

\end{enumerate}

\end{document}